%% file: umwe.tex
\pdfoutput=1
\documentclass[11pt,a4paper]{article}
\usepackage[hyperref]{emnlp2018}
\usepackage{times}
\usepackage{latexsym}
\usepackage{amsmath,amsthm,amssymb}
\usepackage{mathrsfs}
\usepackage{stmaryrd}
\usepackage{graphicx}
\usepackage{subcaption}
\usepackage{booktabs}
\usepackage{algorithm}
\usepackage[noend]{algpseudocode}

\usepackage{url}

\aclfinalcopy 

\DeclareCaptionFont{10pt}{\fontsize{10pt}{11pt}\selectfont}
\newcommand{\ra}[1]{\renewcommand{\arraystretch}{#1}}
\newcommand{\secref}[1]{\S\ref{#1}}

\algnewcommand{\LeftComment}[1]{\State \(\triangleright\) #1}
\newcommand{\pluseq}{\mathrel{+}=}
\newcommand{\expe}{\mathop{{}\mathbb{E}}}

\newcommand{\pivot}{BWE-Pivot}
\newcommand{\direct}{BWE-Direct}
\newcommand{\cd}{\mathcal{D}}
\newcommand{\cm}{\mathcal{M}}
\newcommand{\cl}{\mathscr{L}}
\newcommand{\ce}{\mathcal{E}}
\newcommand{\cv}{\mathcal{V}}
\newcommand{\cs}{\mathcal{S}}
\newcommand{\ct}{\mathcal{T}}

\newcommand{\mat}{\texttt{MAT}}
\newcommand{\mpsr}{\texttt{MPSR}}
\newcommand{\vn}[1]{\vnform{#1}}
\newcommand{\vnform}[1]{\mathtt{#1}}

\title{Unsupervised Multilingual Word Embeddings}

\author{Xilun Chen\\
  Department of Computer Science\\
  Cornell Unversity\\
  Ithaca, NY, 14853, USA\\
  {\tt xlchen@cs.cornell.edu}\\\And
  Claire Cardie\\
  Department of Computer Science\\
  Cornell Unversity\\
  Ithaca, NY, 14853, USA\\
  {\tt cardie@cs.cornell.edu}\\
}

\date{}

\begin{document}
\maketitle

\begin{abstract}
\input{inputs/abstract.tex}
\end{abstract}

\input{inputs/intro.tex}
\input{inputs/relatedwork.tex}
\input{inputs/model.tex}
\input{inputs/experiments.tex}
\input{inputs/conclusion.tex}

\bibliography{umwe}
\bibliographystyle{acl_natbib}

\end{document}

%% file: inputs/abstract.tex
Multilingual Word Embeddings (MWEs) represent words from multiple languages in a single distributional vector space.
Unsupervised MWE (UMWE) methods acquire multilingual embeddings without cross-lingual supervision, which is a significant advantage over traditional supervised approaches and opens many new possibilities for low-resource languages.
Prior art for learning UMWEs, however, merely relies on a number of independently trained Unsupervised Bilingual Word Embeddings (UBWEs) to obtain multilingual embeddings.
These methods fail to leverage the interdependencies that exist among many languages.
To address this shortcoming, we propose a fully unsupervised framework for learning MWEs\footnote{Code: \url{https://github.com/ccsasuke/umwe}} that directly exploits the relations between all language pairs.
Our model substantially outperforms previous approaches in the experiments on multilingual word translation and cross-lingual word similarity.
In addition, our model even beats supervised approaches trained with cross-lingual resources.

%% file: inputs/intro.tex
\section{Introduction}\label{sec:intro}

Continuous distributional word representations~\cite{turian-ratinov-bengio:2010:ACL} have become a common technique across a wide variety of NLP tasks.
Recent research, moreover, proposes cross-lingual word representations~\cite{klementiev-titov-bhattarai:2012:PAPERS,DBLP:journals/corr/MikolovLS13} that create a shared embedding space for words across \emph{two} (Bilingual Word Embeddings, BWE) or \emph{more} languages (Multilingual Word Embeddings, MWE).
Words from different languages with similar meanings will be close to one another in this cross-lingual embedding space.
These embeddings have been found beneficial for a number of cross-lingual and even monolingual NLP tasks~\cite{faruqui-dyer:2014:EACL,DBLP:journals/corr/AmmarMTLDS16}.

The most common form of cross-lingual word representations is the BWE, which connects the lexical semantics of two languages.
Traditionally for training BWEs, cross-lingual supervision is required, either in the form of parallel corpora~\cite{klementiev-titov-bhattarai:2012:PAPERS,zou-EtAl:2013:EMNLP}, or in the form of bilingual lexica~\cite{DBLP:journals/corr/MikolovLS13,xing-EtAl:2015:NAACL-HLT}.
This makes learning BWEs for low-resource language pairs much more difficult.
Fortunately, there are attempts to reduce the dependence on bilingual supervision by requiring a very small parallel lexicon such as identical character strings~\cite{DBLP:journals/corr/SmithTHH17}, or numerals~\cite{artetxe-labaka-agirre:2017:Long}.
Furthermore, recent work proposes approaches to obtain unsupervised BWEs without relying on \emph{any} bilingual resources~\cite{zhang-EtAl:2017:Long5,lample2018word}.

In contrast to BWEs that only focus on a pair of languages, MWEs instead strive to leverage the interdependencies among multiple languages to learn a multilingual embedding space.
MWEs are desirable when dealing with multiple languages simultaneously and have also been shown to improve the performance on some bilingual tasks thanks to its ability to acquire knowledge from other languages~\cite{DBLP:journals/corr/AmmarMTLDS16,duong-EtAl:2017:EACLlong}.
Similar to training BWEs, cross-lingual supervision is typically needed for training MWEs, and the prior art for obtaining \emph{fully unsupervised} MWEs simply maps all the languages independently to the embedding space of a chosen target language\footnote{Henceforth, we refer to this method as \pivot{} as the target language serves as a pivot to connect other languages.} (usually English)~\cite{lample2018word}.
There are downsides, however, when using a single fixed target language with no interaction between any of the two source languages.
For instance, French and Italian are very similar, and the fact that each of them is individually converted to a less similar language, English for example, in order to produce a shared embedding space will inevitably degrade the quality of the MWEs.

For certain multilingual tasks such as translating between any pair of $N$ given languages, another option for obtaining UMWEs exists.
One can directly train UBWEs for each of such language pairs (referred to as \direct{}).
This is seldom used in practice, since it requires training $O(N^2)$ BWE models as opposed to only $O(N)$ in \pivot{}, and is too expensive for most use cases.
Moreover, this method still does not fully exploit the language interdependence.
For example, when learning embeddings between French and Italian, \direct{} only utilizes information from the pair itself, but other Romance languages such as Spanish may also provide valuable information that could improve performance.

In this work, we propose a novel unsupervised algorithm to train MWEs using \emph{only} monolingual corpora (or equivalently, monolingual word embeddings). 
Our method exploits the interdependencies between any two languages and maps all monolingual embeddings into a shared multilingual embedding space via a two-stage algorithm consisting of (i) Multilingual Adversarial Training (\mat{}) and (ii) Multilingual Pseudo-Supervised Refinement (\mpsr{}).
As shown by experimental results on multilingual word translation and cross-lingual word similarity, our model is as efficient as \pivot{} yet outperforms both \pivot{} and \direct{} despite the latter being much more expensive.
In addition, our model achieves a higher overall performance than state-of-the-art \emph{supervised} methods in these experiments.

%% file: inputs/relatedwork.tex
\section{Related Work}\label{sec:relatedwork}
There is a plethora of literature on learning cross-lingual word representations, focusing either on a pair of languages, or multiple languages at the same time~\cite[\textit{inter alia}]{klementiev-titov-bhattarai:2012:PAPERS,zou-EtAl:2013:EMNLP,DBLP:journals/corr/MikolovLS13,45190,coulmance-EtAl:2015:EMNLP,DBLP:journals/corr/AmmarMTLDS16,duong-EtAl:2017:EACLlong}.
One shortcoming of these methods is the dependence on cross-lingual supervision such as parallel corpora or bilingual lexica.
Abundant research efforts have been made to alleviate such dependence~\cite{vulic-moens:2015:ACL-IJCNLP,artetxe-labaka-agirre:2017:Long,DBLP:journals/corr/SmithTHH17}, but consider only the case of a single pair of languages (BWEs).
Furthermore, fully unsupervised methods exist for learning BWEs~\cite{zhang-EtAl:2017:Long5,lample2018word,artetxe2018acl}.
For unsupervised MWEs, however, previous methods merely rely on a number of independent BWEs to separately map each language into the embedding space of a chosen target language~\cite{DBLP:journals/corr/SmithTHH17,lample2018word}.

Adversarial Neural Networks have been successfully applied to various cross-lingual NLP tasks where annotated data is not available, such as cross-lingual text classification~\cite{2016arXiv160601614C}, unsupervised BWE induction~\cite{zhang-EtAl:2017:Long5,lample2018word} and unsupervised machine translation~\cite{lample2018unsupervised,artetxe2018unsupervised}.
These works, however, only consider the case of two languages, and our \mat{} method (\secref{sec:mat}) is a generalization to multiple languages.

\newcite{DBLP:journals/corr/MikolovLS13} first propose to learn cross-lingual word representations by learning a linear mapping between the monolingual embedding spaces of a pair of languages.
It has then been observed that enforcing the linear mapping to be orthogonal could significantly improve performance~\cite{xing-EtAl:2015:NAACL-HLT,artetxe-labaka-agirre:2016:EMNLP2016,DBLP:journals/corr/SmithTHH17}.
These methods solve a linear equation called the orthogonal Procrustes problem for the optimal orthogonal linear mapping between two languages, given a set of word pairs as supervision.
\newcite{artetxe-labaka-agirre:2017:Long} find that when using weak supervision (e.g. digits in both languages), applying this Procrustes process iteratively achieves higher performance.
\newcite{lample2018word} adopt the iterative Procrustes method with pseudo-supervision in a fully unsupervised setting and also obtain good results.
In the MWE task, however, the multilingual mappings no longer have a closed-form solution, and we hence propose the \mpsr{} algorithm (\secref{sec:mpsr}) for learning multilingual embeddings using gradient-based optimization methods.

%% file: inputs/model.tex
\section{Model}\label{sec:model}

In this work, our goal is to learn a single multilingual embedding space for $N$ languages, without relying on \emph{any} cross-lingual supervision.
We assume that we have access to monolingual embeddings for each of the $N$ languages, which can be obtained using unlabeled monolingual corpora~\cite{DBLP:journals/corr/MikolovSCCD13,bojanowski2016enriching}.
We now present our unsupervised MWE (UMWE) model that jointly maps the monolingual embeddings of all $N$ languages into a single space by explicitly leveraging the interdependencies between arbitrary language pairs, but is computationally as efficient as learning $O(N)$ BWEs (instead of $O(N^2)$).

Denote the set of languages as $\cl$ with $|\cl| = N$.
Suppose for each language $l\in \cl$ with vocabulary $\cv_l$, we have a set of $d$-dimensional monolingual word embeddings $\ce_l$ of size $|\cv_l| \times d$.
Let $\cs_l$ denote the monolingual embedding space for $l$, namely the distribution of the monolingual embeddings of $l$.
If a set of embeddings $\ce$ are in an embedding space $\cs$, we write $\ce\vdash\cs$ (e.g. $\forall l: \ce_l\vdash\cs_l$).
Our models learns a set of encoders $\cm_l$, one for each language $l$, and the corresponding decoders $\cm_l^{-1}$.
The encoders map all $\ce_l$ to a single target space $\ct$: $\cm_l(\ce_l) \vdash \ct$.
On the other hand, a decoder $\cm_l^{-1}$ maps an embedding in $\ct$ back to $\cs_l$.

Previous research~\cite{DBLP:journals/corr/MikolovLS13} shows that there is a strong linear correlation between the vector spaces of two languages, and that learning a complex non-linear neural mapping does not yield better results.
\newcite{xing-EtAl:2015:NAACL-HLT} further show that enforcing the linear mappings to be orthogonal matrices achieves higher performance.
Therefore, we let our encoders $\cm_l$ be orthogonal linear matrices, and the corresponding decoders can be obtained by simply taking the transpose: $\cm_l^{-1}=\cm_l^\top$.
Thus, applying the encoder or decoder to an embedding vector is accomplished by multiplying the vector with the encoder/decoder matrix.

Another benefit of using linear encoders and decoders (also referred to as \emph{mappings}) is that we can learn $N-1$ mappings instead of $N$ by choosing the target space $\ct$ to be the embedding space of a specific language (denoted as the \emph{target language}) without losing any expressiveness of the model.
Given a MWE with an arbitrary $\ct$, we can construct an equivalent one with only $N-1$ mappings by multiplying the encoders of each language $\cm_l$ to the decoder of the chosen target language $\cm_t^\top$:
\begin{align*}
     \cm'_t &= \cm_t^\top \cm_t = I\nonumber\\
     \cm'_l \ce_l &= (\cm_t^\top \cm_l) \ce_l \vdash \cs_t
\end{align*}
\noindent
where $I$ is the identity matrix.
The new MWE is isomorphic to the original one.

We now present the two major components of our approach, Multilingual Adversarial Training (\secref{sec:mat}) and Multilingual Pseudo-Supervised Refinement (\secref{sec:mpsr}).

\subsection{Multilingual Adversarial Training}\label{sec:mat}
In this section, we introduce an adversarial training approach for learning multilingual embeddings without cross-lingual supervision.
Adversarial Training is a powerful technique for minimizing the divergence between complex distributions that are otherwise difficult to directly model~\cite{NIPS2014_5423}.
In the cross-lingual setting, it has been successfully applied to unsupervised cross-lingual text classification~\cite{2016arXiv160601614C} and unsupervised bilingual word embedding learning~\cite{zhang-EtAl:2017:Long5,lample2018word}.
However, these methods only consider one pair of languages at a time, and do not fully exploit the cross-lingual relations in the multilingual setting.

\input{inputs/figures/mat_fig.tex}

Figure~\ref{fig:model_adv} shows our Multilingual Adversarial Training (\mat{}) model and the training procedure is described in Algorithm~\ref{alg:mat}.
Note that as explained in \secref{sec:model}, the encoders and decoders adopted in practice are orthogonal linear mappings while the shared embedding space is chosen to be the same space as a selected target language.

In order to learn a multilingual embedding space without supervision, we employ a series of \emph{language discriminators} $\cd_l$, one for each language $l\in\cl$.
Each $\cd_l$ is a binary classifier with a sigmoid layer on top, and is trained to identify how likely a given vector is from $\cs_l$, the embedding space of language $l$.
On the other hand, to train the mappings, we convert a vector from a random language $\vn{lang}_i$ to another random language $\vn{lang}_j$ (via the target space $\ct$ first).
The objective of the mappings is to confuse $\cd_j$, the language discriminator for $\vn{lang}_j$, so the mappings are updated in a way that $\cd_j$ cannot differentiate the converted vectors from the real vectors in $\cs_j$.
This multilingual objective enables us to explicitly exploit the relations between all language pairs during training, leading to improved performance. 
\input{inputs/algorithms/alg_mat.tex}

Formally, for any language $\vn{lang}_j$, the objective that $\cd_j$ is minimizing is:
\begin{equation}
\begin{split}
    J_{\cd_j} =& \expe_{i\sim\cl} \expe_{\substack{x_i\sim\cs_i\\ x_j\sim\cs_j}} \left\llbracket \vphantom{L_d\left(0, \cd_j(\cm_j^\top \cm_i x_i)\right)} L_d\left(1, \cd_j(x_j)\right) + \right.\\
    & \left. L_d\left(0, \cd_j(\cm_j^\top \cm_i x_i)\right) \right\rrbracket
\end{split}
    \label{eqn:j_d}
\end{equation}
\noindent
where $L_d(y, \hat{y})$ is the loss function of $\cd$, which is chosen as the \emph{cross entropy loss} in practice.
$y$ is the language label with $y=1$ indicates a real embedding from that language.

Furthermore, the objective of $\cm_i$ for $\vn{lang}_i$ is:
\begin{equation}
    J_{\cm_i} = \expe_{j\sim\cl} \expe_{\substack{x_i\sim\cs_i\\ x_j\sim\cs_j}} L_d\left(1, \cd_j(\cm_j^\top \cm_i x_i)\right)
    \label{eqn:j_m}
\end{equation}
\noindent
where $\cm_i$ strives to make $\cd_j$ believe that a converted vector to $\vn{lang}_j$ is instead real.
This adversarial relation between $\cm$ and $\cd$ stimulates $\cm$ to learn a shared multilingual embedding space by making the converted vectors look as authentic as possible so that $\cd$ cannot predict whether a vector is a genuine embedding from a certain language or converted from another language via $\cm$.

In addition, we allow $\vn{lang}_i$ and $\vn{lang}_j$ to be the same language in (\ref{eqn:j_d}) and (\ref{eqn:j_m}).
In this case, we are encoding a language to $\ct$ and back to itself, essentially forming an adversarial autoencoder~\cite{makhzani2015adversarial}, which is reported to improve the model performance~\cite{zhang-EtAl:2017:Long5}.
Finally, on Line \ref{alg:line:mat_d_for_loop} and \ref{alg:line:mat_m_for_loop} in Algorithm~\ref{alg:mat}, a for loop is used instead of random sampling.
This is to ensure that in each step, every discriminator (or mapping) is getting updated at least once, so that we do not need to increase the number of training iterations when adding more languages.
Computationally, when compared to the \pivot{} and \direct{} baselines, one step of \mat{} training costs similarly to $N$ BWE training steps, and in practice we train \mat{} for the same number of iterations as training the baselines.
Therefore, \mat{} training scales linearly with the number of languages similar to \pivot{} (instead of quadratically as in \direct{}).

\subsection{Multilingual Pseudo-Supervised Refinement}\label{sec:mpsr}

Using \mat{}, we are able to obtain UMWEs with reasonable quality, but they do not yet achieve state-of-the-art performance.
Previous research on learning unsupervised BWEs~\cite{lample2018word} observes that the embeddings obtained from adversarial training do a good job aligning the frequent words between two languages, but performance degrades when considering the full vocabulary.
They hence propose to use an iterative refinement method~\cite{artetxe-labaka-agirre:2017:Long} to repeatedly refine the embeddings obtained from the adversarial training.
The idea is that we can anchor on the more accurately predicted relations between frequent words to improve the mappings learned by adversarial training.

When learning MWEs, however, it is desirable to go beyond aligning each language with the target space individually, and instead utilize the relations between all languages as we did in \mat{}.
Therefore, we in this section propose a generalization of the existing refinement methods to incorporate a multilingual objective.

\input{inputs/algorithms/alg_mpsr.tex}

In particular, \mat{} can produce an approximately aligned embedding space.
As mentioned earlier, however, the training signals from $\cd$ for rare words are noisier and may lead to worse performance.
Thus, the idea of Multilingual Pseudo-Supervised Refinement (\mpsr{}) is to induce a dictionary of highly confident word pairs for every language pair, used as pseudo supervision to improve the embeddings learned by \mat{}.
For a specific language pair $(\vn{lang}_i, \vn{lang}_j)$, the pseudo-supervised lexicon $\vn{Lex}(\vn{lang}_i, \vn{lang}_j)$ is constructed from \emph{mutual nearest neighbors} between $\cm_i \ce_i$ and $\cm_j \ce_j$, among the most frequent $15k$ words of both languages.

With the constructed lexica, the \mpsr{} objective is:
\begin{equation}
    J_r = \expe_{(i,j)\sim\cl^2} \expe_{(x_i,x_j)\sim \vn{Lex}(i,j)} L_r(\cm_i x_i, \cm_j x_j)
    \label{eqn:j_mpsr}
\end{equation}
\noindent
where $L_r(x, \hat{x})$ is the loss function for $\mpsr{}$, for which we use the \emph{mean square loss}.
The \mpsr{} training is depicted in Algorithm~\ref{alg:mpsr}.

\paragraph{Cross-Lingual Similarity Scaling (CSLS)}
When constructing the pseudo-supervised lexica, a distance metric between embeddings is needed to compute nearest neighbors.
Standard distance metrics such as the Euclidean distance or cosine similarity, however, can lead to the \emph{hubness} problem in high-dimensional spaces when used to calculate nearest neighbors~\cite{Radovanovic:2010:HSP:1756006.1953015,DBLP:journals/corr/DinuB14}.
Namely, some words are very likely to be the nearest neighbors of many others (hubs), while others are not the nearest neighbor of any word.
This problem is addressed in the literature by designing alternative distance metrics, such as the inverted softmax~\cite{DBLP:journals/corr/SmithTHH17} or the CSLS~\cite{lample2018word}.
In this work, we adopt the CSLS similarity as a drop-in replacement for cosine similarity whenever a distance metric is needed.
The CSLS similarity (whose negation is a distance metric) is calculated as follows:
\begin{equation}
\begin{split}
    \text{CSLS}(x, y) &= 2\cos(x, y)\\
    &- \frac{1}{n}\sum_{y'\in N_Y(x)}\cos(x,y')\\
    &- \frac{1}{n}\sum_{x'\in N_X(y)}\cos(x',y)
    \label{eqn:csls}
\end{split}
\end{equation}
\noindent
where $N_Y(x)$ is the set of $n$ nearest neighbors of $x$ in the vector space that $y$ comes from: $Y=\{y_1, ..., y_{|Y|}\}$, and vice versa for $N_X(y)$.
In practice, we use $n=10$.

\subsection{Orthogonalization}\label{sec:ortho}

As mentioned in \secref{sec:model}, orthogonal linear mappings are the preferred choice when learning transformations between the embedding spaces of different languages~\cite{xing-EtAl:2015:NAACL-HLT,DBLP:journals/corr/SmithTHH17}.
Therefore, we perform an orthogonalization update~\cite{pmlr-v70-cisse17a} after each training step to ensure that our mappings $\cm$ are (approximately) orthogonal:
\begin{equation*}
    \forall l: \cm_l = (1+\beta)\cm_l - \beta\cm_l\cm_l^\top\cm_l
\end{equation*}
\noindent where $\beta$ is set to $0.001$.

\subsection{Unsupervised Multilingual Validation}
In order to do model selection in the unsupervised setting, where no validation set can be used, a surrogate validation criterion is required that does not depend on bilingual data.
Previous work shows promising results using such surrogate criteria for model validation in the bilingual case~\cite{lample2018word}, and we in this work adopt a variant adapted to our multilingual setting:
\begin{align*}
    V(\cm, \ce) &= \expe_{(i,j)\sim P_{ij}}  \vn{mean\_csls}(\cm_j^\top\cm_i \ce_i, \ce_j) \nonumber\\
    &= \sum_{i\neq j} p_{ij}\cdot \vn{mean\_csls}(\cm_j^\top\cm_i \ce_i, \ce_j)
\end{align*}
\noindent
where $p_{ij}$ forms a probability simplex.
In this work, we let all $p_{ij}=\frac{1}{N(N-1)}$ so that $V(\cm,\ce)$ reduces to the macro average over all language pairs.
Using different $p_{ij}$ values can place varying weights on different language pairs, which might be desirable in certain scenarios.

The $\vn{mean\_csls}$ function is an unsupervised bilingual validation criterion proposed by~\newcite{lample2018word}, which is the mean CSLS similarities between the most frequent $10k$ words and their translations (nearest neighbors).

%% file: inputs/figures/mat_fig.tex
\begin{figure}
  \centering
  \includegraphics[width=0.35\textwidth]{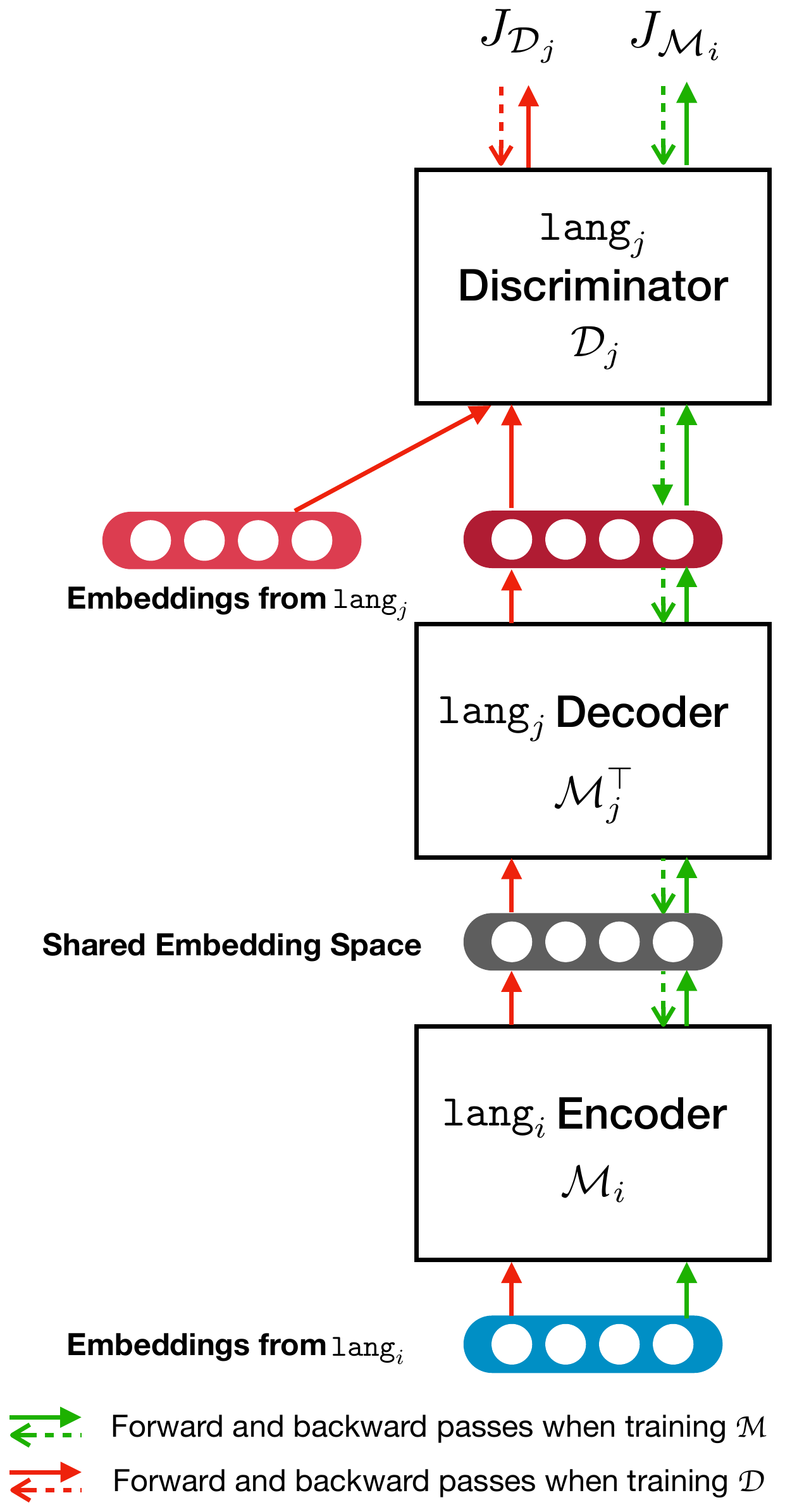}
  \caption{
  Multilingual Adversarial Training (Algorithm~\ref{alg:mat}).
  $\vn{lang}_i$ and $\vn{lang}_j$ are two randomly selected languages at each training step.
  $J_{\cd_j}$ and $J_{\cm_i}$ are the objectives of $\cd_j$ and $\cm_i$, respectively (Eqn.~\ref{eqn:j_d} and \ref{eqn:j_m}).
  }
  \label{fig:model_adv}
\end{figure}

%% file: inputs/algorithms/alg_mat.tex
\begin{algorithm}[t]
\small
\begin{algorithmic}[1]
\Require
Vocabulary $\cv_i$ for each language $\vn{lang}_i\in\cl$.
Hyperparameter $k\in\mathbb{N}$.
\Repeat
\LeftComment{$\cd$ iterations}
\For{$\vn{diter} = 1$ to $k$}
\State $\vn{loss}_d = 0$
\ForAll{$\vn{lang}_j \in \cl$}\label{alg:line:mat_d_for_loop}
\State Select at random $\vn{lang}_i\in \cl$
\State Sample a batch of words $x_i \sim \cv_i$
\State Sample a batch of words $x_j \sim \cv_j$
\State $\hat{x}_t = \cm_i(x_i)$ \Comment{encode to $\ct$}
\State $\hat{x}_j = \cm_j^\top(\hat{x}_t)$ \Comment{decode to $\cs_j$}
\State $y_j = \cd_j(x_j)$ \Comment{real vectors}
\State $\hat{y}_j = \cd_j(\hat{x}_j)$ \Comment{converted vectors}
\State $\vn{loss}_d \pluseq L_d(1, y_j) + L_d(0, \hat{y}_j)$
\EndFor
\State Update all $\cd$ parameters to minimize $\vn{loss}_d$
\EndFor

\LeftComment{$\cm$ iteration}
\State $\vn{loss} = 0$
\ForAll{$\vn{lang}_i \in \cl$}\label{alg:line:mat_m_for_loop}
\State Select at random $\vn{lang}_j\in \cl$
\State Sample a batch of words $x_i \sim \cv_i$
\State $\hat{x}_t = \cm_i(x_i)$ \Comment{encode to $\ct$}
\State $\hat{x}_j = \cm_j^\top(\hat{x}_t)$ \Comment{decode to $\cs_j$}
\State $\hat{y}_j = \cd_j(\hat{x}_j)$
\State $\vn{loss} \pluseq L_d(1, \hat{y}_j)$ 
\EndFor
\State Update all $\cm$ parameters to minimize $\vn{loss}$
\State $\vn{orthogonalize}(\cm)$ \Comment{see \secref{sec:ortho}}
\Until{convergence}

\end{algorithmic}
\caption{Multilingual Adversarial Training}
\label{alg:mat}
\end{algorithm}

%% file: inputs/algorithms/alg_mpsr.tex
\begin{algorithm}[t]
\small
\begin{algorithmic}[1]
\Require
A set of (pseudo-)supervised lexica of word pairs between each pair of languages $\vn{Lex}(\vn{lang}_i, \vn{lang}_j)$.
\Repeat
\State $\vn{loss} = 0$
\ForAll{$\vn{lang}_i \in \cl$}
\State Select at random $\vn{lang}_j\in \cl$
\State Sample $(x_i, x_j) \sim \vn{Lex}(\vn{lang}_i, \vn{lang}_j)$
\State $t_i = \cm_i(x_i)$ \Comment{encode $x_i$}
\State $t_j = \cm_j(x_j)$ \Comment{encode $x_j$}
\State $\vn{loss} \pluseq L_r(t_i, t_j)$ \Comment{refinement loss}
\EndFor
\State Update all $\cm$ parameters to minimize $\vn{loss}$
\State $\vn{orthogonalize}(\cm)$ \Comment{see \secref{sec:ortho}}
\Until{convergence}

\end{algorithmic}
\caption{Multilingual Pseudo-Supervised Refinement}
\label{alg:mpsr}
\end{algorithm}

%% file: inputs/experiments.tex
\section{Experiments}\label{sec:experiments}

\input{inputs/tables/tab_wmt}

In this section, we present experimental results to demonstrate the effectiveness of our unsupervised MWE method on two benchmark tasks, the multilingual word translation task, and the SemEval-2017 cross-lingual word similarity task.
We compare our \mat{}+\mpsr{} method with state-of-the-art unsupervised and supervised approaches, and show that ours outperforms previous methods, supervised or not, on both tasks.

Pre-trained $300d$ fastText (monolingual) embeddings~\cite{bojanowski2016enriching} trained on the Wikipedia corpus are used for all systems that require monolingual word embeddings for learning cross-lingual embeddings.

\subsection{Multilingual Word Translation}\label{sec:exp_wt}

In this section, we consider the task of word translation between arbitrary pairs of a set of $N$ languages.
To this end, we use the recently released multilingual word translation dataset on six languages: English, French, German, Italian, Portuguese and Spanish~\cite{lample2018word}.
For any pair of the six languages, a ground-truth bilingual dictionary is provided with a train-test split of $5000$ and $1500$ unique source words, respectively.
The $5k$ training pairs are used in training supervised baseline methods, while all unsupervised methods do not rely on any cross-lingual resources.
All systems are tested on the $1500$ test word pairs for each pair of languages.

For comparison, we adopted a state-of-the-art unsupervised BWE method~\cite{lample2018word} and generalize it for the multilingual setting using the two aforementioned approaches, namely \pivot{} and \direct{}, to produce unsupervised baseline MWE systems.
English is chosen as the pivot language in \pivot{}.
We further incorporate the supervised \direct{} (Sup-\direct{}) method as a baseline, where each BWE is trained on the $5k$ gold-standard word pairs via the orthogonal Procrustes process~\cite{artetxe-labaka-agirre:2017:Long,lample2018word}.

\begin{table*}[t]
\small
\centering
\ra{1.2}
\begin{tabular}{lccccccccccc}
    \toprule
    & en-de & en-es & de-es & en-it & de-it & es-it & en-fa & de-fa & es-fa & it-fa & \textbf{Average} \\
    \midrule
    \multicolumn{12}{l}{\emph{Supervised methods with cross-lingual supervision}}\\
    Luminoso & \bf .769 & \bf .772 & \bf .735 & \bf .787 & \bf .747 & \bf .767 & .595 & .587 & .634 & .606 & .700 \\
    NASARI & .594 & .630 & .548 & .647 & .557 & .592 & .492 & .452 & .466 & .475 & .545 \\
    \midrule
    \multicolumn{12}{l}{\emph{Unsupervised methods without cross-lingual supervision}}\\
    \pivot{} & .709 & .711 & .703 & .709 & .682 & .721 & .672 & .655 & .701 & .688 & .695 \\
    \direct{} & .709 & .711 & .703 & .709 & .675 & .726 & .672 & .662 & .714 & .695 & .698 \\
    \midrule
    \mat{}+\mpsr{} & .711 & .712 & .708 & .709 & .684 & .730 & \bf .680 & \bf .674 & \bf .720 & \bf .709 & \bf .704 \\
    \bottomrule
\end{tabular}
\caption{Results for the SemEval-2017 Cross-Lingual Word Similarity task.
Spearman's $\rho$ is reported.
Luminoso~\cite{DBLP:conf/semeval/SpeerL17} and NASARI~\cite{CAMACHOCOLLADOS201636} are the two top-performing systems for SemEval-2017 that reported results on all language pairs.
}
\label{tab:semeval}
\end{table*}

Table~\ref{tab:mwt} presents the evaluation results, wherein the numbers represent \emph{precision@1}, namely how many times one of the correct translations of a source word is retrieved as the top candidate. 
All systems retrieve word translations using the CSLS similarity in the learned embedding space.
Table~\ref{tab:mwt-a} shows the detailed results for all $30$ language pairs,
while Table~\ref{tab:mwt-b} summarizes the results in a number of ways.
We first observe the training cost of all systems summarized in Table~\ref{tab:mwt-b}.
\#BWEs indicates the training cost of a certain method measured by how many BWE models it is equivalent to train.
\pivot{} needs to train $2(N{-}1)$ BWEs since a separate BWE is trained for each direction in a language pair for increased performance.
\direct{} on the other hand, trains an individual BWE for all (again, directed) pairs, resulting a total of $N(N{-}1)$ BWEs.
The supervised Sup-\direct{} method trains the same number of BWEs as \direct{} but is much faster in practice, for it does not require the unsupervised adversarial training stage.
Finally, while our \mat{}+\mpsr{} method does not train independent BWEs, as argued in \secref{sec:mat}, the training cost is roughly equivalent to training $N{-}1$ BWEs, which is corroborated by the real training time shown in Table~\ref{tab:mwt-b}.

We can see in Table~\ref{tab:mwt-a} that our \mat{}+\mpsr{} method achieves the highest performance on all but 3 language pairs, compared against both the unsupervised and supervised approaches.
When looking at the overall performance across all language pairs, \direct{} achieves a $+0.6\%$ performance gain over \pivot{} at the cost of being much slower to train.
When supervision is available, Sup-\direct{} further improves another $0.4\%$ over \direct{}.
Our \mat{}+\mpsr{} method, however, attains an impressive $1.3\%$ improvement against Sup-\direct{}, despite the lack of cross-lingual supervision.

To provide a more in-depth examination of the results, we first consider the Romance language pairs, such as fr-es, fr-it, fr-pt, es-it, it-pt and their reverse directions.
\pivot{} performs notably worse than \direct{} on these pairs, which validates our hypothesis that going through a less similar language (English) when translating between similar languages will result in reduced accuracy.
Our \mat{}+\mpsr{} method, however, overcomes this disadvantage of \pivot{} and achieves the best performance on all these pairs through an explicit multilingual learning mechanism without increasing the computational cost.

Furthermore, our method also beats the \direct{} approach, which supports our second hypothesis that utilizing knowledge from languages beyond the pair itself could improve performance.
For instance, there are a few pairs where \pivot{} outperforms \direct{}, such as de-it, it-de and pt-de, even though it goes through a third language (English) in \pivot{}.
This might suggest that for some less similar language pairs, leveraging a third language as a bridge could in some cases work better than only relying on the language pair itself.
German is involved in all these language pairs where \pivot{} outperforms than \direct{}, which is potentially due to the similarity between German and the pivot language English.
We speculate that if choosing a different pivot language, there might be other pairs that could benefit.
This observation serves as a possible explanation of the superior performance of our multilingual method over \direct{}, since our method utilizes knowledge from all languages during training.

\subsection{Cross-Lingual Word Similarity}\label{sec:exp_ws}

In this section, we evaluate the quality of our MWEs on the cross-lingual word similarity (CLWS) task, which assesses how well the similarity in the cross-lingual embedding space corresponds to a human-annotated semantic similarity score.
The high-quality CLWS dataset from SemEval-2017~\cite{camachocollados-EtAl:2017:SemEval} is used for evaluation.
The dataset contains word pairs from any two of the five languages: English, German, Spanish, Italian, and Farsi (Persian), annotated with semantic similarity scores.

In addition to the \pivot{} and \direct{} baseline methods, we also include the two best-performing systems on SemEval-2017, Luminoso~\cite{DBLP:conf/semeval/SpeerL17} and NASARI~\cite{CAMACHOCOLLADOS201636} for comparison.
Note that these two methods are supervised, and have access to the Europarl\footnote{\url{http://opus.nlpl.eu/Europarl.php}} (for all languages but Farsi) and the OpenSubtitles2016\footnote{\url{http://opus.nlpl.eu/OpenSubtitles2016.php}} parallel corpora.

Table~\ref{tab:semeval} shows the results, where the performance of each model is measured by the Spearman correlation.
When compared to the \pivot{} and the \direct{} baselines, \mat{}+\mpsr{} continues to perform the best on all language pairs.
The qualitative findings stay the same as in the word translation task, except the margin is less significant.
This might be because the CLWS task is much more lenient compared to the word translation task, where in the latter one needs to correctly identify the translation of a word out of hundreds of thousands of words in the vocabulary.
In CLWS though, one can still achieve relatively high correlation in spite of minor inaccuracies.

On the other hand, an encouraging result is that when compared to the state-of-the-art supervised results, our \mat{}+\mpsr{} method outperforms NASARI by a very large margin, and achieves top-notch overall performance similar to the competition winner, Luminoso, without using any bitexts.
A closer examination reveals that our unsupervised method lags a few points behind Luminoso on the European languages wherein the supervised methods have access to the large-scale high-quality Europarl parallel corpora.
It is the low-resource language, Farsi, that makes our unsupervised method stand out.
All of the unsupervised methods outperform the supervised systems from SemEval-2017 on language pairs involving Farsi, which is not covered by the Europarl bitexts.
This suggests the advantage of learning unsupervised embeddings for lower-resourced languages, where the supervision might be noisy or absent.
Furthermore, within the unsupervised methods, \mat{}+\mpsr{} again performs the best, and attains a higher margin over the baseline approaches on the low-resource language pairs, vindicating our claim of better multilingual performance.

%% file: inputs/tables/tab_wmt.tex
\begin{table*}[t]
\small
\ra{1.2}
    \centering
    \begin{subtable}{\textwidth}
    \centering
    \begin{tabular}{@{\hspace{0.45em}}l@{\hspace{0.45em}}@{\hspace{0.45em}}c@{\hspace{0.45em}}@{\hspace{0.45em}}c@{\hspace{0.45em}}@{\hspace{0.45em}}c@{\hspace{0.45em}}@{\hspace{0.45em}}c@{\hspace{0.45em}}@{\hspace{0.45em}}c@{\hspace{0.45em}}@{\hspace{0.45em}}c@{\hspace{0.45em}}@{\hspace{0.45em}}c@{\hspace{0.45em}}@{\hspace{0.45em}}c@{\hspace{0.45em}}@{\hspace{0.45em}}c@{\hspace{0.45em}}@{\hspace{0.45em}}c@{\hspace{0.45em}}@{\hspace{0.45em}}c@{\hspace{0.45em}}@{\hspace{0.45em}}c@{\hspace{0.45em}}@{\hspace{0.45em}}c@{\hspace{0.45em}}@{\hspace{0.45em}}c@{\hspace{0.45em}}@{\hspace{0.45em}}c@{\hspace{0.45em}}}
        \toprule
            & en-de & en-fr & en-es & en-it & en-pt & de-fr & de-es & de-it & de-pt & fr-es & fr-it & fr-pt & es-it & es-pt & it-pt \\
        \midrule
        \multicolumn{16}{l}{\emph{Supervised methods with cross-lingual supervision}}\\
        Sup-\direct{} & 73.5 & 81.1 & 81.4 & 77.3 & 79.9 & 73.3 & 67.7 & 69.5 & 59.1 & 82.6 & 83.2 & 78.1 & 83.5 & 87.3 & 81.0 \\
        \midrule
        \multicolumn{16}{l}{\emph{Unsupervised methods without cross-lingual supervision}}\\
        \pivot{} & 74.0 & 82.3 & 81.7 & 77.0 & 80.7 & 71.9 & 66.1 & 68.0 & 57.4 & 81.1 & 79.7 & 74.7 & 81.9 & 85.0 & 78.9 \\
        \direct{} & 74.0 & 82.3 & 81.7 & 77.0 & 80.7 & 73.0 & 65.7 & 66.5 & 58.5 & 83.1 & 83.0 & 77.9 & 83.3 & 87.3 & 80.5\\
        \midrule
        \mat{}+\mpsr{} & \bf 74.8 & \bf 82.4 & \bf 82.5 & \bf 78.8 & \bf 81.5 & \bf 76.7 & \bf 69.6 & \bf 72.0 & \bf 63.2 & \bf 83.9 & \bf 83.5 & \bf 79.3 & \bf 84.5 & \bf 87.8 & \bf 82.3 \\
        \specialrule{\heavyrulewidth}{\aboverulesep}{\belowrulesep}
            & de-en & fr-en & es-en & it-en & pt-en & fr-de & es-de & it-de & pt-de & es-fr & it-fr & pt-fr & it-es & pt-es & pt-it \\
        \midrule
        \multicolumn{16}{l}{\emph{Supervised methods with cross-lingual supervision}}\\
        Sup-\direct{} & 72.4 & \bf 82.4 & 82.9 & 76.9 & \bf 80.3 & 69.5 & 68.3 & 67.5 & 63.7 & 85.8 & 87.1 & 84.3 & 87.3 & 91.5 & 81.1 \\
        \midrule
        \multicolumn{16}{l}{\emph{Unsupervised methods without cross-lingual supervision}}\\
        \pivot{} & 72.2 & 82.1 & 83.3 & \bf 77.7 & 80.1 & 68.1 & 67.9 & 66.1 & 63.1 & 84.7 & 86.5 & 82.6 & 85.8 & 91.3 & 79.2 \\
        \direct{} & 72.2 & 82.1 & 83.3 & \bf 77.7 & 80.1 & 69.7 & 68.8 & 62.5 & 60.5 & 86 & 87.6 & 83.9 & 87.7 & 92.1 & 80.6 \\
        \midrule
        \mat{}+\mpsr{} &\bf 72.9 & 81.8 & \bf 83.7 & 77.4 & 79.9 & \bf 71.2 & \bf 69.0 & \bf 69.5 & \bf 65.7 & \bf 86.9 & \bf 88.1 & \bf 86.3 & \bf 88.2 & \bf 92.7 & \bf 82.6 \\
        \bottomrule
    \end{tabular}
    \vspace{-1em}
    \caption{Detailed Results}
    \label{tab:mwt-a}
    \end{subtable}
    
    \vspace{1em}
    \begin{subtable}{\textwidth}
    \centering
    \begin{tabular}{@{\hspace{0.27em}}l@{\hspace{0.27em}}@{\hspace{0.27em}}c@{\hspace{0.27em}}@{\hspace{0.27em}}c@{\hspace{0.27em}}c@{\hspace{0.27em}}c@{\hspace{0.27em}}@{\hspace{0.27em}}c@{\hspace{0.27em}}@{\hspace{0.27em}}c@{\hspace{0.27em}}@{\hspace{0.27em}}c@{\hspace{0.27em}}@{\hspace{0.27em}}c@{\hspace{0.27em}}@{\hspace{0.27em}}c@{\hspace{0.27em}}c@{\hspace{0.27em}}c@{\hspace{0.27em}}@{\hspace{0.27em}}c@{\hspace{0.27em}}@{\hspace{0.27em}}c@{\hspace{0.27em}}@{\hspace{0.27em}}c@{\hspace{0.27em}}@{\hspace{0.27em}}c@{\hspace{0.27em}}@{\hspace{0.27em}}c@{\hspace{0.27em}}@{\hspace{0.27em}}c@{\hspace{0.27em}}}
        \toprule
        & \multicolumn{2}{c}{\emph{Training Cost}} && \multicolumn{6}{c}{\emph{Single Source}} && \multicolumn{6}{c}{\emph{Single Target}} &  \\
        \cmidrule{2-3}\cmidrule{5-10}\cmidrule(r){12-17}
        & \#BWEs & time && en-xx & de-xx & fr-xx & es-xx & it-xx & pt-xx && xx-en & xx-de & xx-fr & xx-es & xx-it & xx-pt & \textbf{Overall} \\
        \midrule
        \multicolumn{18}{l}{\emph{Supervised methods with cross-lingual supervision}}\\
        Sup-\direct{} & $N(N{-}1)$ & $4h$ && 78.6 & 68.4 & 79.2 & 81.6 & 80.0 & 80.2 && 79.0 & 68.5 & 82.3 & 82.1 & 78.9 & 77.1 & 78.0\\
        \midrule
        \multicolumn{18}{l}{\emph{Unsupervised methods without cross-lingual supervision}}\\
        \pivot{} & $2(N{-}1)$ & $8h$ && 79.1 & 67.1 & 77.1 & 80.6 & 79.0 & 79.3 && \bf 79.1 & 67.8 & 81.6 & 81.2 & 77.2 & 75.3 & 77.0\\
        \direct{} & $N(N{-}1)$ & $23h$ && 79.1 & 67.2 & 79.2 & 81.7 & 79.2 & 79.4 && \bf 79.1 & 67.1 & 82.6 & 82.1 & 78.1 & 77.0 & 77.6\\
        \midrule
        \mat{}+\mpsr{} & $N{-}1$ & $5h$ && \bf 80.0 & \bf 70.9 & \bf 79.9 & \bf 82.4 & \bf 81.1 & \bf 81.4 && \bf 79.1 & \bf 70.0 & \bf 84.1 & \bf 83.4 & \bf 80.3 & \bf 78.8 & \bf 79.3\\
        \bottomrule
    \end{tabular}
    \vspace{-1em}
    \caption{Summarized Results}
    \label{tab:mwt-b}
    \end{subtable}
    \caption{Multilingual Word Translation Results for English, German, French, Spanish, Italian and Portuguese.
    The reported numbers are \emph{precision@1} in percentage.
    All systems use the nearest neighbor under the CSLS distance for predicting the translation of a certain word.
    }
    \label{tab:mwt}
\end{table*}

%% file: inputs/conclusion.tex
\section{Conclusion}\label{sec:conclusion}
In this work, we propose a fully unsupervised model for learning multilingual word embeddings (MWEs).
Although methods exist for learning high-quality unsupervised BWEs~\cite{lample2018word}, little work has been done in the unsupervised multilingual setting.
Previous work relies solely on a number of unsupervised BWE models to generate MWEs (e.g. \pivot{} and \direct{}), which does not fully leverage the interdependencies among all the languages.
Therefore, we propose the \mat{}+\mpsr{} method that explicitly exploits the relations between all language pairs without increasing the computational cost.
In our experiments on multilingual word translation and cross-lingual word similarity (SemEval-2017), we show that \mat{}+\mpsr{} outperforms existing unsupervised and even supervised models, achieving new state-of-the-art performance.

For future work, we plan to investigate how our method can be extended to work with other BWE frameworks, in order to overcome the instability issue of~\newcite{lample2018word}.
As pointed out by recent work~\cite{P18-1072,artetxe2018acl}, the method by~\newcite{lample2018word} performs much worse on certain languages such as Finnish, etc.
More reliable multilingual embeddings might be obtained on these languages if we adapt our multilingual training framework to work with the more robust methods proposed recently.